\documentclass{article}

\usepackage{placeins}
\usepackage{pdflscape}
\usepackage{multirow}

\usepackage[dblblindworkshop, final]{neurips_2026}
\workshoptitle{TS-LIMITS}

\usepackage[utf8]{inputenc} 
\usepackage[T1]{fontenc}    
\usepackage{hyperref}       
\usepackage{url}            
\usepackage{booktabs}       
\usepackage{amsfonts}       
\usepackage{nicefrac}       
\usepackage{microtype}      
\usepackage{xcolor}         

\usepackage{graphicx}
\usepackage{amsmath}
\usepackage{mathtools}
\mathtoolsset{showonlyrefs}

\title{STeP-TS: Shared Temporal Filters with Pruning for Time Series}

\author{%
  Hanne Raum\thanks{raum@informatik.uni-freiburg.de} \\
  Department of Computer Science\\
  Albert-Ludwigs-Universität Freiburg\\
  \texttt{raum@informatik.uni-freiburg.de} \\
  \And
  Jasper Hoffmann\\
  Department of Computer Science\\
  Albert-Ludwigs-Universität Freiburg\\
  \AND
  Joschka Bödecker\\
  Department of Computer Science and BrainLinks-BrainTools\\
  Albert-Ludwigs-Universität Freiburg\\}

\begin{document}

\maketitle

\begin{abstract}
Multivariate time series forecasting requires capturing temporal dependencies across multiple scales while accounting for both shared and variable-specific dynamics. Compact forecasting models have emerged as an attractive alternative to increasingly large architectures, but preserving strong predictive performance within tight computational and memory budgets remains challenging.
We propose a compact one-layer convolutional framework that learns a shared bank of multi-scale temporal filters of each variable. Lightweight variable-specific scalar weights determine the contribution of each temporal component, enabling the model to capture shared temporal structure while adapting to variable-specific dynamics. The learned weights and filter-output energy further provide a natural signal for structured pruning of convolutional filters, reducing computational complexity with minimal degradation in forecasting performance. We additionally introduce an amplitude-adaptive normalization mechanism that preserves variable-dependent temporal amplitudes.
Experiments on multiple datasets demonstrate forecasting performance competitive with state-of-the-art methods while requiring substantially fewer parameters and inference FLOPs and achieving low per-sample latency.
\end{abstract}

\section{Introduction}

Forecasting is crucial in many real-world applications such as weather prediction \citep{weather}, transportation systems \citep{traffic_flow}, and power demand estimation \citep{alsalem_hybrid_2025}. Time-series forecasting has evolved from classical statistical models such as ARIMA \citep{ARIMA} to deep-learning approaches including LSTMs \citep{LSTM}, convolutional models such as ModernTCN \citep{luo_moderntcn_2024}, and Transformers such as PatchTST \citep{nie_time_2022_Patch_TST}.

However, increasing model complexity does not necessarily improve forecasting performance. DLinear \citep{zeng_are_2023_dlinear}, for example, showed that simple linear mappings can outperform substantially larger Transformer models on several long-term forecasting benchmarks. This raises the question of how much model capacity is actually required for accurate forecasting, especially in resource-constrained settings.

\begin{figure}[htb]
  \centering
  \includegraphics[width=1\linewidth]{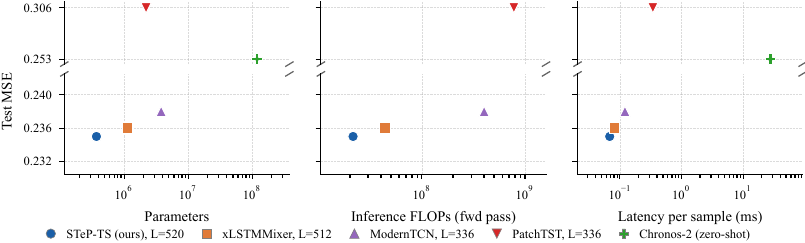}
  \caption{Accuracy against cost on Weather (horizon = 336). STeP-TS matches the strongest baselines in test MSE (0.235 vs 0.236 / 0.238) while using 3.1× fewer parameters and 3.0× fewer inference FLOPs than xLSTMMixer, and 10.2× / 27.3× fewer than ModernTCN. Each model is measured at its own published configuration, so lookback differs per model (given in the legend). PatchTST uses default hyperparameters, as additional reference Chronos2 \citep{chronos2} (zero-shot) is added}
  \label{intro:advertisment}
\end{figure}

A similar question arises for convolutional architectures. Convolutions can be interpreted as learnable filters whose frequency response is determined by their kernel weights, while the kernel size influences the temporal receptive field. Temporal Convolutional Networks (TCNs) increase the receptive field through dilated convolutions (\citep{TCN_1}, \citep{bai_empirical_2018_TCN_2}), while more recent architectures such as ModernTCN employ large-kernel convolutions to efficiently capture long-range temporal dependencies \citep{luo_moderntcn_2024}. This motivates investigating how kernel size, receptive field, and model complexity influence forecasting performance, and how much convolutional capacity is necessary to capture the relevant temporal structure.

In this work, we propose \textbf{STeP-TS} (Shared Temporal Filters with Pruning for Time Series), a simple one-layer model that decomposes the input into parallel multi-scale temporal components using convolutional filters with different kernel sizes. As shown in Fig.~\ref{intro:advertisment}, STeP-TS achieves competitive forecasting accuracy while requiring substantially fewer parameters and inference FLOPs than strong baselines.

Our design is motivated by the observation that variables in multivariate time series often exhibit shared temporal and spectral characteristics, while retaining variable-specific dynamics and interactions. We therefore use shared temporal filters across all variables, an assumption also adopted by approaches such as \citep{chen_tsmixer_2023,das_long-term_2024_TiDE,kraus_xlstm-mixer_2025}. Rather than learning separate filters for each variate learned scalar weights combine the resulting components to adapt the shared filters to variable-specific dynamics. 
Furthermore, the learned weights, together with the energy of the convolutional outputs, provide a simple criterion for identifying and pruning entire convolutional filters, reducing computational complexity without additional optimization or performance-based pruning. The shallow architecture makes the contribution of individual filters directly observable, providing a transparent basis for structured pruning.

We make the following contribution: 
\begin{enumerate}
    \item We propose a simple one-layer convolutional architecture that achieves competitive state-of-the-art forecasting performance while requiring substantially less architectural complexity
    \item We improve computational efficiency through  a simple, filter-pruning mechanism that uses the learned feature-specific weights and the energy of the convolutional filter outputs to identify and remove entire convolutional filters without significant performance degradation.
    \item We mitigate the damping effect of convolutional filters by introducing a learned amplitude-adaptive normalization mechanism that preserves and adapts the amplitude of temporal patterns across variables.
\end{enumerate}

\section{Related work}
In the following, we provide a short overview of different architectures used in time series forecasting.

\textbf{Recurrent Neural Networks (RNNs)}, particularly Long Short-Term Memory (LSTM) \citep{LSTM}, model temporal dependencies recurrently. XLSTMTime \citep{alharthi_xlstmtime_2024} extends LSTM with modified memory structures and exponential gating, while XLSTM-Mixer \citep{kraus_xlstm-mixer_2025} combines this approach with an NLinear-based backbone \citep{zeng_are_2023_dlinear}. \textbf{State Space Models (SSMs)} provide another approach, with SMamba \citep{wang_is_2025_Smamba} adapting Mamba to forecasting through bidirectional and additional linear layers.

\textbf{MLP-based models} include DLinear \citep{zeng_are_2023_dlinear}, which decomposes series into trend and seasonal components, as well as TiDE \citep{das_long-term_2024_TiDE} and CycleNet \citep{lin_cyclenet_2024}, which incorporate covariates and cyclic patterns, respectively. TSMixer \cite{chen_tsmixer_2023} mixes information across temporal and feature dimensions, while TimeMixer \citep{wang_timemixer_2024} adopts a multiscale decomposition and mixing strategy inspired by filter banks \citep{filterBanks} and wavelet-based multiresolution analysis \citep{mallat_MRA}.

\textbf{Transformer-based models} such as Informer \citep{Informer}, Autoformer \citep{Autoformer}, and FEDformer \citep{Fedformer} introduced increasingly complex architectures, although their performance was challenged by simpler models such as DLinear. PatchTST \citep{nie_time_2022_Patch_TST} improved upon DLinear using patch-based Transformers with variable-independent processing. More recently, pretrained time-series foundation models such as Chronos-2 \citep{ansari2025chronos2} and Moirai \citep{liu2026moirai2} have scaled forecasting through large-scale pretraining, at substantially higher computational cost.

\textbf{Convolutional Neural Networks (CNNs)} capture temporal patterns through convolutional operations. SCINet \citep{SCINet} uses recursive decomposition, MICN \citep{wang2023micn} combines seasonal-trend decomposition with multiscale pooling, and ModernTCN \citep{luo_moderntcn_2024} employs patch embedding and large-kernel depthwise convolutions.

In contrast, our approach avoids explicit decomposition, patchification, and hierarchical processing. It learns parallel temporal filters with different kernel sizes, providing multiple receptive fields, while per-variable and per-feature weighting enables variable-specific adaptation. Filters can be pruned to adapt model capacity, and large strides with interpolation enable efficient large-kernel filtering. The resulting model forms a compact and potentially sparse multi-scale temporal filter bank.

\section{Method}
\begin{figure}[t]
  \centering
  \includegraphics[width=1.\linewidth]{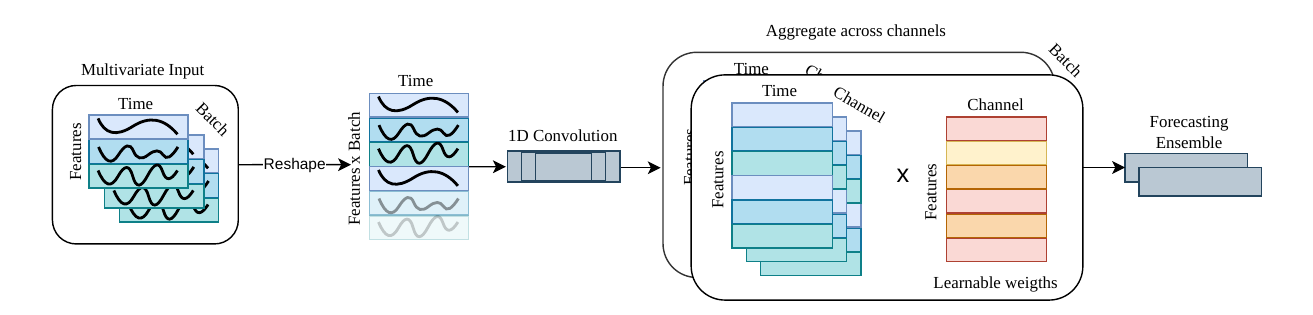}
  \caption{\textbf{Overview of STeP-TS.} The input features are first reshaped into the batch dimension and processed by 1D convolutional filters with different kernel sizes along the sequence length, learning general temporal patterns across the dataset. The resulting representations are reshaped into $\text{batch} \times \text{features} \times \text{channels}$, where channels are individually weighted for each feature and then aggregated. The resulting $\text{batch} \times \text{features} \times \text{sequence}$ representation is finally passed to the forecasting head to generate the prediction.}
  \label{method:fig:overview}
\end{figure}

In the following, we describe the model architecture, an overview is given in Fig.~\ref{method:fig:overview}.  For consistency, we define the notation for multivariate forecasting. The input is a time series vector $\mathbf{X}_T = \{\mathbf{x}_{1},\dots,\mathbf{x}_{T}\} \in \mathbb{R}^{V \times T}$, where $T$ denotes the number of time steps and $V$ the number of variables. The goal is to predict the future time sequence $\mathbf{Y}_H = \{\mathbf{x}_{T+1},\dots,\mathbf{x}_{T+H}\} \in \mathbb{R}^{V \times H}$ over a forecasting horizon of length $H$.

\textbf{Decomposition module.}
The decomposition module assumes that variates within the same domain share common temporal patterns, e.g. due to similar sampling rates or underlying periodicities. Instead of learning independent filters for each variate, the model learns shared temporal filters across all variates to capture common frequency patterns efficiently.

Given an input time series $\mathbf{X}_{T} \in \mathbb{R}^{B \times V \times T } $, we merge the batch $B$ and the variate dimension to $ \mathbf{\tilde{X}} \in \mathbb{R}^{(B \cdot V) \times T }$. A set of $C$ one-dimensional convolutional filters $\{k_c \}_{c=1}^C$ with different kernel sizes is applied along the temporal axis,
\begin{equation}
    \mathbf{\tilde{X}}^\mathrm{conv}= k_c * \mathbf{\tilde{X}} , \; \; c=1,...,C,
\end{equation}
enforcing the learning of filters that generalize across both samples and features.
or pruning, the convolutional weights are normalized as described in Appendix \ref{app: unitnorm}. 
After convolution, the representation is reshaped back to $\mathbf{X}^\mathrm{conv} \in \mathbb{R}^{B \times V \times C \times T }$.
To retain variate-specific characteristics, learnable aggregation weights  $\mathbf{A_{v,c}} \in \mathbb{R}^{V \times C } $ determine the contribution of each channel to each variate:
\begin{equation} \label{aggregate_conv}
     \mathbf{X}^\mathrm{out}_{b,v,t} = \sum_{c=1}^C \mathbf{A_{v,c}}\mathbf{X}^\mathrm{conv}_{b,v,c,t}
\end{equation}
resulting into an final output $ \mathbf{X}^\mathrm{out} \in \mathbb{R}^{B \times V \times T }$, which are passed to an ensemble head for the final forecast. Additional the aggregation weights are later used to provide a signal for pruning unnecessary convolutional channels. 
To reduce the cost of large kernels, convolution is performed with increased stride and the resulting representations are interpolated to the original temporal resolution, details in Appendix \ref{interpolate_large_kernel}.

\paragraph{Ensemble forecasting head.}
After the convolutional stack, each variate $v$ is represented by $x_v\in\mathbb{R}^T$. A single shared forecasting head may not capture differences between variates, while using one head per variate can lead to overfitting. We therefore use a small number $K=1,2,3$ of shared linear heads $h_{v,k} \in \mathbb{R}^{H\times T}$ and let each variate mix heads using its own weights $\Pi_{v,k}$:
\begin{equation}
\hat y_v=\sum_{k=1}^{K}\Pi_{v,k}h_{v,k},
\qquad
\text{with } \Pi=\operatorname{softmax}_{K}\bigl(UV^\top/\sqrt r\bigr).
\end{equation}

where the mixing weights $\Pi_{v,k}$ are generated from low rank factors $U\in\mathbb{R}^{V\times r}$, and $V\in\mathbb{R}^{K\times r}$.
The mixture weights are fixed for each variate and independent on the current input. With ($K=1$), all variates use the same head. Larger ($K$) gives them more freedom to specialize. Although with $K\ge V$ we could in principle give each variate its own head. We only consider $K=1,2,3$ to keep the model compact and let a hyperparameter study choose its best fit for each dataset.

\subsection{Normalization}
To improve forecasting performance, we use reversible instance normalization (RevIN) \citep{Kim2022ReversibleIN} to account for distribution shifts between input sequences, details in Appendix \ref{app:RevIn}. But implicitly assuming compatible input and horizon distributions. This can lead to amplitude mismatch under fast regime changes, such as weekday to weekend transitions, an instance of \emph{inter-space} distribution shift \citep{fan2023dishts}. We therefore introduce a bounded, input-conditioned gain that adjusts the RevIN denormalization scale (adRevIN) while remaining shared across variates and the forecast horizon. The gain is predicted from a coarse representation of the raw input using a small MLP, it starts from the RevIN solution through zero initialization, and is learned directly from the forecasting loss without auxiliary statistics targets, more details in Appendix \ref{app:regime_aware amplitude correction}.

\subsection{Structured pruning of convolutional channels} \label{meth:pruning}
Since the channels are aggregated using the learnable weights $\mathbf{A_{v,c}}$, setting an  entire column $\mathbf{A}_{:,c}$ to zero removes the contribution of channel $c$ for all variates. Pruning therefore reduces to selecting a subset of the $C$ convolutional channels.

\paragraph{Importance.}
The importance of a channel depends on both its learned aggregation weights and the temporal activity of its convolutional response. Let $g_c(x)_{v,t}$ denote the output of convolutional channel $c$ for variate $v$ at time $t$ with $N_b$ as samples per batch and $B$ as batches, 
\begin{equation}
  \sigma^{2}_{v,c}(x) \;=\; \frac{1}{T-1}\sum_{t=1}^{T}
    \bigl(g_c(x)_{v,t} - \bar{g}_{v,c}(x)\bigr)^{2},
  \qquad
  \bar{g}_{v,c}(x) = \frac{1}{T}\sum_{t=1}^{T} g_c(x)_{v,t}.
  \label{eq:tvar}
\end{equation}
The variance is preferred over the temporal mean because oscillatory signals can have means close to zero despite exhibiting substantial temporal variation. We then define the importance score $p_c^{\mathrm{e}}$ as 
\begin{equation}
p_c^{\mathrm{e}} = \frac{1}{N_b B V}\sum_{n=1}^{N_b}\sum_{b=1}^{B}\sum_{v=1}^{V}
      a_{v,c}^{2}\,\sigma^{2}_{v,c}(x_{n,b})
\end{equation}
Since $\mathbf{A_{v,c}}$ is constant over time, $\sigma^{2}_{v,c}(x)$ gives an additional estimate over the temporal variance of the output. $p_c^{\mathrm{e}}$ accounts for both. 

\paragraph{Pruning ratio.}
With the importance score $p_c^{\mathrm{e}}$ the effective number of useful scales is estimated using the participation ratio of channels $C$,
\begin{equation}
\mathrm{PR}(p)
=
\frac{\left(\sum_c p_c^{\mathrm{e}}\right)^2}
{\sum_c (p_c^{\mathrm{e}})^2}
=
\frac{1}{\sum_c s_c^2},
\qquad
\text{with the importance score }
s_c =
\frac{p_c}{\sum_j p_j}.
\label{eq}
\end{equation}
The participation ratio lies between $1$ and $C$ and represents the effective number of scales contributing to the representation, we set the pruning floor round up $\mathrm{PR}(p)$ and set a margin $m=1$ channels.  
Recomputing $\mathrm{PR}(p)$ after each removal would progressively reduce the pruning target. If a channel with importance share $u$ is removed, the participation ratio of the remaining channels becomes
\begin{equation}
\mathrm{PR}'
=
\mathrm{PR}
\frac{(1-u)^2}
{1-\mathrm{PR}u^2}.
\end{equation}
For the least important channel, $u\leq1/C$, which implies $\mathrm{PR}'<\mathrm{PR}$. Thus, recomputing the participation ratio would allow further pruning purely due to the changing normalisation. We therefore compute the initial participation ratio once and fix the pruning budget to
$
\left\lceil \mathrm{PR}_0 \right\rceil + m
$
for the remainder of training, while the importance score $s_c$ and therefore the rank of the channels is updated after every pruned channel. Details of the pruning schedule can be found in the Appendix ~\ref{pruning_schedule}.

\section{Experiments}
We conduct experiments on standard long-term forecasting tasks. Furthermore, we perform an ablation study.
\paragraph{Datasets.} For long term forecasting we used 6 established benchmarks (ETTh1, ETTm1, ETTh2, ETTm2, weather, traffic, electricity), details in the Appendix \ref{exp:details}. 
\paragraph{Training.} We selected the parameters for STeP-TS by Hyperparamter studies with Optuna\citep{Optuna} and Tree Parzen Estimator(TPE)\citep{TPE}. We  trained on mean absoulute error (MAE) and evaluated results on mean squared errors (MSE) and MAE (except traffic). The final results are averaged over three different seeds (2021, 2022, 2023), following \citep{wang_timemixer_2024}). All models were trained on a single NVIDIA L40S on an internal SLURM cluster (PyTorch 2.6, CUDA 12.4), optimised with Adam under a cosine schedule with linear warmup, and tuned with Optuna 4.3. 
\paragraph{Baseline Models.} We compare STeP-TS against a wide range of state-of-the-art baselines, including convolutional architectures such ModernTCN \citep{luo_moderntcn_2024}, Transformer-based models such as PatchTST \citep{nie_time_2022_Patch_TST} and, recurrent architectures including xLSTMTime \citep{alharthi_xlstmtime_2024} and xLSTM-Mixer \citep{kraus_xlstm-mixer_2025} TimeMixer \citep{wang_timemixer_2024} and TS Mixer \citep{alharthi_xlstmtime_2024}, state-space models including S-Mamba \citep{wang_is_2025_Smamba}, MLP-based approaches such as DLinear \citep{zeng_are_2023_dlinear}, CycleNet \citep{lin_cyclenet_2024}, and TiDE \citep{das_long-term_2024_TiDE}, as well as pretrained models including Timer-XL \citep{liu_timer-xl_2025} and $\text{Morai}_{\text{Base}}$ \citep{woo_unified_2024-morai}.
For completeness, all baselines are reported in the Appendix; however, to improve readability, we present only a representative subset in the main text and refer otherwise to Table~\ref{app:fullresults}.

\subsection{Main Results: Long-Term Forecasting Performance}
Table \ref{res:aveg_long} presents the long-term forecasting results. We evaluate STeP-TS with a varying lookback lenghth sequence lengths of 520 or 720, following the argumentation in \citep{kraus_xlstm-mixer_2025, abdelmalak_channel_2025_lookback}. HPO selects between the two configurations. The corresponding hyperparameter configurations are reported in Table \ref{tab:hpo_hyperparameters}, while the complete results for each forecasting horizon are provided in Appendix \ref{app:fullresults}.

\begin{table}[!ht]
    \centering
    \setlength{\tabcolsep}{1pt}
    \scriptsize
    \caption{Average (\textit{Avg.}) results across four prediction horizons {96, 192, 336, 720}. Lower MSE and MAE indicate better performance. Best results are highlighted in bold red, while second-best results are shown in blue and underlined. All results are averaged over three random seeds (2021, 2022, 2023); average standard deviation of 0.001 for both MSE and MAE.}
    \label{res:aveg_long}
    \resizebox{0.95\textwidth}{!}{%
    \begin{tabular}{l|cc|cc|cc|cc|cc|cc|cc|cc|cc|cc}
    \toprule

    \textbf{Model Type}
    & \multicolumn{4}{c}{\textbf{Convolutional}}
    & \multicolumn{4}{c}{\textbf{RNN}}
    & \multicolumn{6}{c}{\textbf{MLP}}
    & \multicolumn{2}{c}{\textbf{SSM}}
    & \multicolumn{4}{c}{\textbf{Transformer}}
    \\

    \cmidrule(lr){2-5}
    \cmidrule(lr){6-9}
    \cmidrule(lr){10-15}
    \cmidrule(lr){16-17}
    \cmidrule(lr){18-21}

    \textbf{Models}
    & \multicolumn{2}{c}{\shortstack{\textbf{STeP-TS}\\ (ours)}}
    & \multicolumn{2}{c}{\shortstack{\textbf{ModTCN} \\ \citeyear{luo_moderntcn_2024}}}
    & \multicolumn{2}{c}{\shortstack{\textbf{xLSTMix.} \\ \citeyear{kraus_xlstm-mixer_2025}}}
    & \multicolumn{2}{c}{\shortstack{\textbf{xLSTMTime} \\ \citeyear{alharthi_xlstmtime_2024}}}
    & \multicolumn{2}{c}{\shortstack{\textbf{TimeMix} \\ \citeyear{wang_timemixer_2024}}}
    & \multicolumn{2}{c}{\shortstack{\textbf{DLinear} \\ \citeyear{zeng_are_2023_dlinear}}}
    & \multicolumn{2}{c}{\shortstack{\textbf{TiDE} \\ \citeyear{das_long-term_2024_TiDE}}}
    & \multicolumn{2}{c}{\shortstack{\textbf{S-Mamba} \\ \citeyear{wang_is_2025_Smamba}}}
    & \multicolumn{2}{c}{\shortstack{\textbf{PatchTST} \\ \citeyear{nie_time_2022_Patch_TST}}}
    & \multicolumn{2}{c}{\shortstack{\textbf{Timer-XL} \\ \citeyear{liu_timer-xl_2025}}}
    \\

    \cmidrule(lr){2-3} \cmidrule(lr){4-5} \cmidrule(lr){6-7}
    \cmidrule(lr){8-9} \cmidrule(lr){10-11} \cmidrule(lr){12-13}
    \cmidrule(lr){14-15} \cmidrule(lr){16-17} \cmidrule(lr){18-19}
    \cmidrule(lr){20-21}

    \textbf{Dataset}
    & MSE & MAE
    & MSE & MAE
    & MSE & MAE
    & MSE & MAE
    & MSE & MAE
    & MSE & MAE
    & MSE & MAE
    & MSE & MAE
    & MSE & MAE
    & MSE & MAE
    \\

\midrule

    Weather
    & \textcolor{red}{\textbf{0.216}} & \textcolor{red}{\textbf{0.246}}
    & 0.224 & 0.264
    & \textcolor{blue}{\underline{0.219}} & \textcolor{blue}{\underline{0.250}}
    & 0.222 & 0.255
    & 0.222 & 0.262
    & 0.246 & 0.300
    & 0.236 & 0.282
    & 0.251 & 0.276
    & 0.241 & 0.264
    & 0.256 & 0.294 \\

    ETTh1
    & 0.407 & 0.420
    & \textcolor{blue}{\underline{0.404}} & 0.420
    & \textcolor{red}{\textbf{0.397}} & 0.420
    & 0.408 & 0.428
    & 0.411 & 0.423
    & 0.423 & 0.437
    & 0.419 & 0.430
    & 0.455 & 0.450
    & 0.458 & 0.450
    & \textcolor{blue}{\underline{0.404}} & \textcolor{blue}{\underline{0.417}} \\

    ETTm1
    & 0.353 & \textcolor{blue}{\underline{0.369}}
    & 0.351 & 0.381
    & \textcolor{red}{\textbf{0.339}} & \textcolor{red}{\textbf{0.366}}
    & \textcolor{blue}{\underline{0.347}} & 0.372
    & 0.348 & 0.376
    & 0.357 & 0.379
    & 0.355 & 0.378
    & 0.398 & 0.405
    & 0.353 & 0.382
    & 0.373 & 0.392 \\

    ETTm2
    & \textcolor{blue}{\underline{0.249}} & \textcolor{red}{\textbf{0.306}}
    & 0.253 & 0.314
    & \textcolor{red}{\textbf{0.248}} & \textcolor{blue}{\underline{0.307}}
    & 0.254 & 0.310
    & 0.256 & 0.316
    & 0.267 & 0.332
    & 0.249 & 0.312
    & 0.288 & 0.332
    & 0.256 & 0.317
    & 0.273 & 0.336 \\

    ETTh2
    & 0.340 & 0.384
    & \textcolor{blue}{\underline{0.322}} & \textcolor{red}{\textbf{0.379}}
    & 0.340 & \textcolor{blue}{\underline{0.382}}
    & 0.346 & 0.386
    & \textcolor{red}{\textbf{0.316}} & 0.384
    & 0.431 & 0.447
    & 0.345 & 0.394
    & 0.381 & 0.405
    & 0.324 & 0.381
    & 0.347 & 0.388 \\

    Electricity
    & 0.160 & \textcolor{blue}{\underline{0.247}}
    & \textcolor{blue}{\underline{0.156}} & 0.253
    & \textcolor{red}{\textbf{0.153}} & \textcolor{red}{\textbf{0.245}}
    & 0.157 & 0.250
    & \textcolor{blue}{\underline{0.156}} & \textcolor{blue}{\underline{0.247}}
    & 0.166 & 0.264
    & 0.159 & 0.257
    & 0.170 & 0.265
    & 0.159 & 0.253
    & 0.193 & 0.295 \\

    Traffic
    & 0.414 & 0.271
    & 0.396 & 0.270
    & 0.392 & 0.253
    & 0.391 & 0.261
    & 0.388 & 0.263
    & 0.434 & 0.295
    & 0.356 & 0.261
    & 0.414 & 0.276
    & 0.391 & 0.264
    & -- & -- \\

\bottomrule
    \end{tabular}
}%
\end{table}

STeP-TS achieves competitive forecasting performance with substantially fewer parameters and lower latency, demonstrating its computational efficiency. In particular, it outperforms ModernTCN on Weather and ETTm2, and on ETTm1 in terms of MAE, while achieving comparable performance on ETTh1. On Electricity, STeP-TS achieves better MAE with only a slight degradation in MSE. These results are particularly notable given the substantially smaller parameter count of STeP-TS, suggesting that the proposed architecture makes more effective use of its parameters.

\subsection{Main Results: Efficiency and Pruning}
\begin{figure}[htb]
\centering
\includegraphics[width=1.\linewidth]{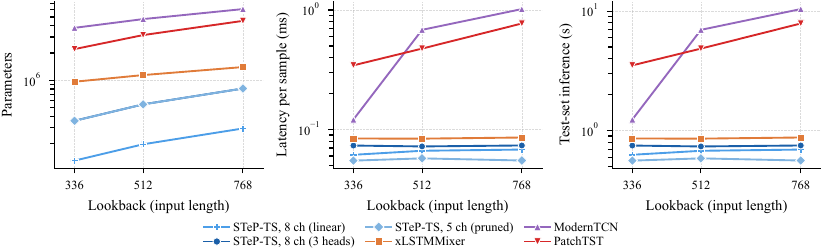}
\caption{Computational cost of each model on Weather at $H=336$. All models are evaluated at the same three lookback windows. Left: parameter count. Centre: per-sample latency, measured as the batch-32 inference time amortised over the 32 series. Right: wall-clock inference time over the full Weather test split at batch size 32. STeP-TS is evaluated at both ends of its head range—the linear head and the published three-head configuration—as well as with a pruned variant that removes three of its eight scale channels (one low-, one mid-, and one high-scale channel) while otherwise remaining identical to the three-head model. All measurements were performed on a single RTX 3080 Ti with a batch size of 32. PatchTST is evaluated its default hyperparameters.}
\label{res}
\end{figure}

STeP-TS achieves competitive performance while maintaining a low computational cost, see Fig.~\ref{intro:advertisment}. Additionally, Fig.~\ref{res} compares the parameter count, inference latency, and overall test-time cost of the evaluated models on Weather for a forecast horizon of 336 and 21 variates. To provide a consistent comparison, all models are evaluated with the same three lookback windows. For STeP-TS, we evaluate both the smallest and largest configurations in terms of the number of heads (one and three heads), as well as different lookback windows.

As described in Sec.~\ref{meth:pruning}, our already compact model can be further pruned by removing entire convolutional layers. Although this has only a minor effect on the total parameter count, it reduces the number of FLOPs and inference latency, resulting in a measurable speed-up. We measure this effect consistently across the ETT and Weather datasets. Details of the speed-up measurements are provided in Table~\ref{tab:effi}.

\subsection{Ablations: Ensemble Head and Normalization}
To investigate the contribution of the different model components to forecasting performance, we train several variants of STeP-TS. Specifically, we compare a single linear forecasting head with an ensemble of two forecasting heads, and investigate the impact of the additional RevIN module (adRevIN) with a scaling factor of $s_c \in [0.1, 0.5]$ and pooling windows of [6,12,24,32, 48, 64], details in \ref{app:regime_aware amplitude correction}. The results are reported in Table \ref{tab:res:ablation}. We perform hyperparameter optimization (HPO) only for the learning rate to identify suitable settings for each variant.
For comparability, we fix the main architectural parameters across all variants: the number of channels, sequence length (520), number of convolutional layers (8), and kernel sizes (30--160). This setup may not yield optimal configurations for every variant, but ensures consistent and directly comparable results. The ablation study is conducted on the Weather, ETTm2, and ETTm1 datasets.

\begin{table}[!ht]
\centering
\scriptsize
\setlength{\tabcolsep}{1pt}
\caption{Ablation study of the ensemble forecasting head and adRevIN across different forecasting horizons with a fixed lookback window of 520. Results are averaged across three seeds (2021, 2022, 2023); with average standard deviation of 0.001 for both MSE and MAE.}
\label{tab:res:ablation}
\resizebox{0.95\columnwidth}{!}{%
\begin{tabular}{l|cccc|cccc|cccc|cccc|cccc}
\toprule
\textbf{Weather}
& \multicolumn{4}{c|}{96}
& \multicolumn{4}{c|}{192}
& \multicolumn{4}{c|}{336}
& \multicolumn{4}{c|}{720}
& \multicolumn{4}{c}{Avg} \\
\cmidrule(lr){2-5}
\cmidrule(lr){6-9}
\cmidrule(lr){10-13}
\cmidrule(lr){14-17}
\cmidrule(lr){18-21}
& \multicolumn{2}{c|}{-adRevIN}
& \multicolumn{2}{c|}{+adRevIN}
& \multicolumn{2}{c|}{-adRevIN}
& \multicolumn{2}{c|}{+adRevIN}
& \multicolumn{2}{c|}{-adRevIN}
& \multicolumn{2}{c|}{+adRevIN}
& \multicolumn{2}{c|}{-adRevIN}
& \multicolumn{2}{c|}{+adRevIN}
& \multicolumn{2}{c|}{-adRevIN}
& \multicolumn{2}{c}{+adRevIN} \\
& MSE & MAE & MSE & MAE
& MSE & MAE & MSE & MAE
& MSE & MAE & MSE & MAE
& MSE & MAE & MSE & MAE
& MSE & MAE & MSE & MAE \\
\midrule

\textbf{With Ensemble}
& \textcolor{red}{\textbf{0.142}} & \textcolor{red}{\textbf{0.181}}
& \textcolor{red}{\textbf{0.142}} & \textcolor{red}{\textbf{0.181}}
& 0.184 & \textcolor{red}{\textbf{0.223}}
& \textcolor{red}{\textbf{0.183}} & \textcolor{red}{\textbf{0.223}}
& \textcolor{red}{\textbf{0.235}} & 0.265
& \textcolor{red}{\textbf{0.235}} & \textcolor{red}{\textbf{0.264}}
& \textcolor{red}{\textbf{0.309}} & \textcolor{red}{\textbf{0.320}}
& 0.310 & \textcolor{red}{\textbf{0.320}}
& \textcolor{red}{\textbf{0.218}} & \textcolor{red}{\textbf{0.247}}
& \textcolor{red}{\textbf{0.218}} & \textcolor{red}{\textbf{0.247}} \\

\textbf{W/o Ensemble}
& 0.145 & 0.187
& 0.145 & 0.189
& 0.188 & 0.229
& 0.188 & 0.229
& 0.238 & 0.268
& 0.240 & 0.271
& 0.312 & 0.323
& 0.314 & 0.324
& 0.221 & 0.252
& 0.222 & 0.253 \\

\midrule
\textbf{ETTm1}
& \multicolumn{4}{c|}{96}
& \multicolumn{4}{c|}{192}
& \multicolumn{4}{c|}{336}
& \multicolumn{4}{c|}{720}
& \multicolumn{4}{c}{Avg} \\
\cmidrule(lr){2-5}
\cmidrule(lr){6-9}
\cmidrule(lr){10-13}
\cmidrule(lr){14-17}
\cmidrule(lr){18-21}
& \multicolumn{2}{c|}{-adRevIN}
& \multicolumn{2}{c|}{+adRevIN}
& \multicolumn{2}{c|}{-adRevIN}
& \multicolumn{2}{c|}{+adRevIN}
& \multicolumn{2}{c|}{-adRevIN}
& \multicolumn{2}{c|}{+adRevIN}
& \multicolumn{2}{c|}{-adRevIN}
& \multicolumn{2}{c|}{+adRevIN}
& \multicolumn{2}{c|}{-adRevIN}
& \multicolumn{2}{c}{+adRevIN} \\
& MSE & MAE & MSE & MAE
& MSE & MAE & MSE & MAE
& MSE & MAE & MSE & MAE
& MSE & MAE & MSE & MAE
& MSE & MAE & MSE & MAE \\
\midrule

\textbf{With Ensemble}
& 0.299 & 0.342
& 0.297 & 0.339
& 0.339 & 0.359
& 0.338 & 0.361
& 0.372 & 0.378
& 0.363 & 0.377
& 0.425 & 0.410
& 0.416 & 0.409
& 0.359 & 0.372
& 0.354 & 0.372 \\

\textbf{W/o Ensemble}
& 0.290 & 0.334
& \textcolor{red}{\textbf{0.285}} & \textcolor{red}{\textbf{0.333}}
& 0.334 & 0.358
& \textcolor{red}{\textbf{0.330}} & \textcolor{red}{\textbf{0.357}}
& 0.363 & \textcolor{red}{\textbf{0.376}}
& \textcolor{red}{\textbf{0.359}} & \textcolor{red}{\textbf{0.376}}
& 0.422 & 0.409
& \textcolor{red}{\textbf{0.410}} & \textcolor{red}{\textbf{0.407}}
& 0.352 & 0.369
& \textcolor{red}{\textbf{0.346}} & \textcolor{red}{\textbf{0.368}} \\

\midrule
\textbf{ETTm2}
& \multicolumn{4}{c|}{96}
& \multicolumn{4}{c|}{192}
& \multicolumn{4}{c|}{336}
& \multicolumn{4}{c|}{720}
& \multicolumn{4}{c}{Avg} \\
\cmidrule(lr){2-5}
\cmidrule(lr){6-9}
\cmidrule(lr){10-13}
\cmidrule(lr){14-17}
\cmidrule(lr){18-21}
& \multicolumn{2}{c|}{-adRevIN}
& \multicolumn{2}{c|}{+adRevIN}
& \multicolumn{2}{c|}{-adRevIN}
& \multicolumn{2}{c|}{+adRevIN}
& \multicolumn{2}{c|}{-adRevIN}
& \multicolumn{2}{c|}{+adRevIN}
& \multicolumn{2}{c|}{-adRevIN}
& \multicolumn{2}{c|}{+adRevIN}
& \multicolumn{2}{c|}{-adRevIN}
& \multicolumn{2}{c}{+adRevIN} \\
& MSE & MAE & MSE & MAE
& MSE & MAE & MSE & MAE
& MSE & MAE & MSE & MAE
& MSE & MAE & MSE & MAE
& MSE & MAE & MSE & MAE \\
\midrule

\textbf{With Ensemble}
& \textcolor{red}{\textbf{0.160}} & \textcolor{red}{\textbf{0.243}}
& 0.164 & 0.248
& \textcolor{red}{\textbf{0.217}} & \textcolor{red}{\textbf{0.283}}
& 0.220 & 0.287
& 0.273 & 0.322
& 0.272 & \textcolor{red}{\textbf{0.321}}
& 0.362 & 0.379
& 0.361 & 0.378
& \textcolor{red}{\textbf{0.253}} & \textcolor{red}{\textbf{0.307}}
& 0.254 & 0.309 \\

\textbf{W/o Ensemble}
& 0.161 & 0.245
& 0.163 & 0.247
& 0.219 & 0.285
& 0.220 & 0.286
& \textcolor{red}{\textbf{0.271}} & \textcolor{red}{\textbf{0.321}}
& \textcolor{red}{\textbf{0.271}} & \textcolor{red}{\textbf{0.321}}
& 0.360 & 0.378
& \textcolor{red}{\textbf{0.359}} & \textcolor{red}{\textbf{0.377}}
& \textcolor{red}{\textbf{0.253}} & \textcolor{red}{\textbf{0.307}}
& \textcolor{red}{\textbf{0.253}} & 0.308 \\

\bottomrule
\end{tabular}
}
\label{tab:ablation_ensemble}
\end{table}

The ablation study in Table \ref{tab:ablation_ensemble} shows that the benefits of the ensemble forecasting head are strongly data-dependent. While Weather consistently benefits from the ensemble, ETTm1 does not show a clear advantage. For ETTm2, the ensemble provides improvements mainly at the shorter forecasting horizons (96 and 192). Based on these dataset-dependent effects, we therefore leave the choice of using the ensemble forecasting head to HPO.
The effect of adRevIN is also dataset-dependent. For ETTm1, adRevIN consistently improves forecasting performance, whereas its impact on Weather is negligible. This is a favorable property, as applying adRevIN does not negatively affect performance when it does not provide a measurable benefit. A similar observation can be made for ETTm2, where adRevIN leads to a slight decrease in MAE while having virtually no effect on MSE. Overall, the ablation suggests that the choice of the ensemble head is the more influential design decision, while adRevIN can provide additional benefits for some datasets without noticeably harming performance when its effect is limited.

\section{Conclusion and Outlook}
In this work, we introduced STeP-TS, a simple and lightweight model for multivariate time-series forecasting. By sharing convolutional filters across variables while learning feature-specific weights, STeP-TS captures multi-scale temporal patterns with a few parameter than existing approaches. Despite its simplicity, STeP-TS achieves competitive, and in several cases state-of-the-art, forecasting performance while offering lower computational cost and latency. Furthermore, the shared structure enables entire convolutional layers to be pruned, providing an additional reduction in computational complexity without requiring additional optimization procedures.
Looking ahead, the compact and structured nature of STeP-TS opens interesting opportunities for interpretability. Since the model relies on a small number of shared temporal components and explicit feature-specific weights, the contribution of individual temporal filters to each variable can be directly analyzed. Combined with the proposed pruning mechanism, this could provide a simple way to identify which temporal patterns are relevant for individual variables. Exploring these interpretability aspects, as well as extending the approach to different temporal structures, represents a promising direction for future work.

\paragraph{Acknowledgments}
 Funded by the Deutsche Forschungsgemeinschaft (DFG, German Research Foundation), Project-ID 499552394 SFB 1597 and  under grant number 539134284, through EFRE (FEIH\_2698644) and the state of Baden\-Württemberg. \includegraphics[width=0.1\textwidth]{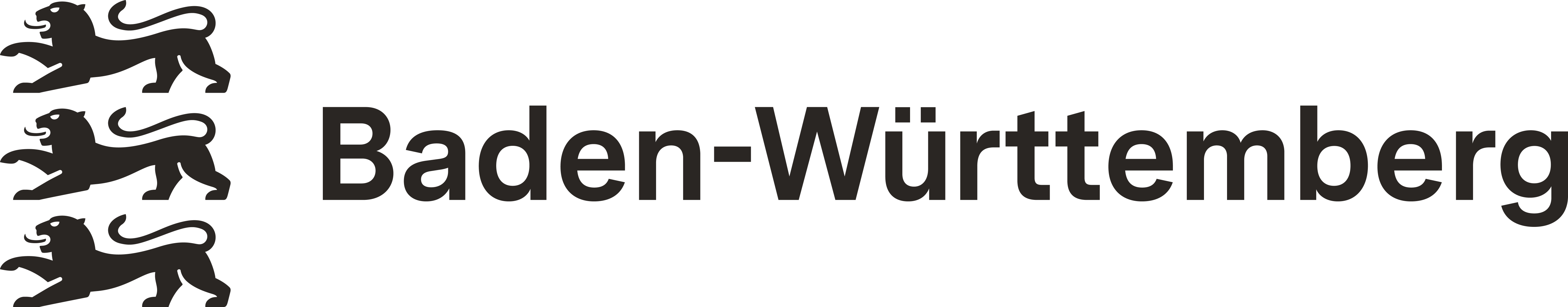}
\includegraphics[width=0.1\textwidth]{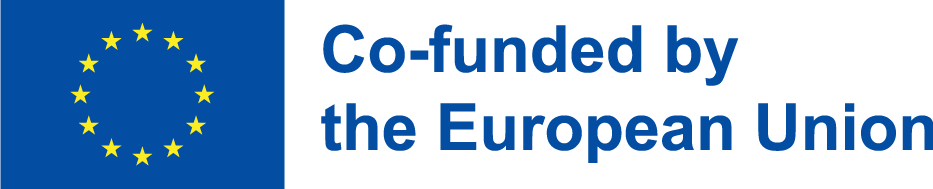}. Use of infrastructure by the Institute for Machine-Brain Interfacing Technology (IMBIT), University of Freiburg, under the Research Buildings at Universities program (project BW1241025). 
\newpage
\FloatBarrier
\bibliographystyle{plainnat}
\bibliography{references}
\newpage

\appendix



\section{Experimental Details}
In the following, we provide definitions such as metrics and norms, normalization schemes, and details on pruning used for the experiments in this work.

\subsection{Unit Norm}\label{app: unitnorm}
With convolutional filters of different kernel sizes, the model can learn a filter bank covering multiple temporal scales. Depending on the dataset, however, not all scales may be necessary. Since the convolutional filters operate in parallel, individual scales can be removed without affecting the information flow through the remaining filters.

To identify redundant scales, we first normalise the convolutional weights,
\begin{equation}
\tilde{w}{:,c}
={\frac{w_{:,c}|}{\lVert w_{:,c} \rVert_2 + \varepsilon}},
\end{equation}
so that filters with different kernel sizes can be compared on an equal basis. Without this normalization, filters with larger kernel sizes would have a larger overall weight magnitude even when their individual weights have the same magnitude. The convolutional scales are then pruned according to the following criteria. \

\subsection{Interpolation of larger convolutional layer} \label{interpolate_large_kernel}
To obtain a consistent output size across all convolutional channels, the convolutional outputs must be restored to the original input length. This becomes computationally expensive for large kernel sizes, as we use kernels of up to $K_c=160$.
Although dilated convolutions could reduce the computational cost, we opted not to use them in order to preserve as much local information as possible. Specifically, the stride determines the spacing between consecutive positions at which the convolution is evaluated. Although not every input position is directly used as the center of a convolution operation, positions between two evaluated points still contribute indirectly through the convolutional window. After the strided convolution, the resulting feature map is upsampled to the original sequence length using linear interpolation. Thus, the values at positions omitted by the strided convolution are reconstructed by interpolating between neighboring convolutional outputs.
The stride $s_c$ is selected based on the kernel size $K_c$ as follows 
\begin{equation}
  s_c = \begin{cases} 1 & K_c < 50, \\ 2 & 50 \le K_c < 100, \\ 4 & K_c \ge 100, \end{cases},
\end{equation}

and the output is interpolated back to length L with 
\begin{equation}
  g_c(x)_{v,t} \;=\; \Bigl(\operatorname{Up}_{s_c}\!\bigl[\,
    \textstyle\sum_{k=0}^{K_c-1} w^{(c)}_k\, \tilde{x}_{v,\, s_c\tau + k} + b_c
  \,\bigr]\Bigr)_{t}. 
\end{equation}

where ($\operatorname{Up}{s_c}$) denotes linear interpolation by a factor corresponding to the stride ($s_c$). This procedure preserves the original temporal resolution while substantially reducing the number of convolution operations for larger kernels.

\subsection{Reversible Instance Normalization (RevIN)} \label{app:RevIn}
We use Reversible Instance Normalization (RevIN) \citep{Kim2022ReversibleIN}, to account for distribution shifts between input sequences. Each sequence $x$ is normalized using its own mean and standard deviation and denormalized after processing, 
\begin{equation}
\tilde{x} = \frac{x-\mu_x}{\sigma_x},
\qquad
\hat{x} = \tilde{x} \; \sigma_x+\mu_x .
\end{equation}
This removes shifts in the input distribution while restoring the original scale at the output. Here we use a simplified version of RevIN. 

\subsection{Regime-aware amplitude correction} \label{app:regime_aware amplitude correction}
RevIN normalizes each input window using its own mean and standard deviation and reuses these statistics to denormalize the forecast. This implicitly assumes that the input and output windows have compatible distributions. While this assumption can be reasonable for slowly varying series, it can break under fast amplitude changes, such as weekday--weekend transitions, where the forecast horizon may have a systematically larger or smaller scale than the lookback window. In this case, the input-derived standard deviation used by RevIN may be inappropriate for the forecast horizon, resulting in an amplitude mismatch between the normalized prediction and the target. This is an instance of the \emph{inter-space} distribution shift described by \citep{fan2023dishts}.

We therefore introduce a single bounded gain $\gamma(x)$ into the RevIN denormalisation:
\begin{equation}
\hat{x} \;=\; \gamma(x)\,\sigma \, \tilde{x} \;+\; \mu_f
\end{equation}
Here, $\gamma(x)$ is computed by
\begin{equation}
  \gamma(x) \;=\; 1 + s\,\tanh\bigl(\phi(x)\bigr) \;\in\; [\,1-s,\;1+s\,],
  \label{eq:ampscale}
\end{equation}
with  $s \in [0.1, 0.5]$ bounds the maximum deviation from RevIn. 
$\phi(x)$ is two-layer MLP from a coarse summary $g(x)$ of the raw lookback window, without RevIN normalization. We obtain $g(x)$ by average-pooling over variates and local time blocks, with pooling stride $p\in{6,12,32,84}$ treated as a hyperparameter. Additional the second MLP layer is initialized to zero, so that $\gamma(x)=1$ and training starts exactly from RevIN \citep{bachlechner2021rezero}.
The gain is shared across variates and the forecast horizon, so it corrects only the overall amplitude across the dataset without changing the forecast shape. Since $\sigma$ is detached, $\gamma$ is the only path through which the forecasting loss can adjust the predicted amplitude.

\paragraph{Pruning schedule.} \label{pruning_schedule}
Pruning is performed during training rather than only after convergence. After a warm-up period, channels are removed at regular intervals separated by a cooldown of $E_{\mathrm{r}}$ epochs. No pruning is performed within the final $E_{\mathrm{r}}$ epochs, ensuring that every removal is followed by sufficient retraining.

Pruned channels are re-zeroed after each optimiser step to prevent gradients or optimiser momentum from reviving them. A forward-pass mask can additionally skip inactive branches, reducing the computational cost without changing the output. Finally, the selected architecture can be retrained from random initialisation, allowing different architectures to be compared independently of a particular optimisation trajectory, but experiments showed that on average performance decreased less within the iterative pruning than retraining the selected architectures from scratch, see ~\ref{tab:energy_vs_random_channels}.

\paragraph{Pruning measurement.} \label{app:speed-Up}
We define the speedup as the ratio between the wall-clock latency of the dense model and that of the pruned model:
\begin{equation}
\mathrm{SpeedUp} = \frac{\mathrm{Latency}_{\mathrm{dense}}}{\mathrm{Latency}_{\mathrm{pruned}}}.
\end{equation}
A speedup of $1.4\times$ therefore indicates a 1.4-fold reduction in inference time, while $1.0\times$ indicates no improvement. Latency is measured for a single forward pass with a batch size of one, averaged over 50 runs following 10 warm-up runs. GPU synchronization is performed for each measurement to ensure that the reported latency reflects actual device execution time rather than an estimate based on FLOPs.
 
\subsection{Experimental Details} \label{exp:details}
\paragraph{Datasets.} \label{app:Datasets}
Table \ref{app:data} provides an overview of the datasets. The table summarizes the key properties of the datasets, including their domains, forecasting horizons, sampling frequencies, and number of variates.

\begin{table}[htb] \label{app:data}
    \centering
    \caption{The long-term forecasting benchmark datasets}
    \label{tab:datasets}
    \begin{tabular}{l l l c c c}
        \toprule
        Dataset & Source & Domain & Horizons & Sampling & Variates \\
        \midrule
        Weather     & \citep{Informer} & Weather        & 96--720 & 10 min      & 21  \\
        Electricity & \citep{Informer} & Power Usage    & 96--720 & 1 hour      & 321 \\
        Traffic     & \citep{Autoformer}   & Traffic Load   & 96--720 & 1 hour      & 862 \\
        ETT         & \citep{Informer} & Power Production & 96--720 & 15\& 60 min & 7   \\
        \bottomrule
    \end{tabular}
\end{table} 
We follow the standard data splits established in \citep{Autoformer}, dividing each dataset chronologically into training, validation, and test sets. Specifically, we use a split ratio of 6:2:2 for the ETT datasets and 7:1:2 for all other datasets

\paragraph{Metric details.}
For long-term forecasting, we use mean squared error (MSE), mean absolute
error (MAE), and mean reverse Huber (BerHu) loss. The MAE and MSE are
defined as
\begin{equation}
    \mathrm{MAE}
    =
    \frac{1}{HV}
    \sum_{i=T+1}^{T+H}
    \sum_{v=1}^{V}
    |y_{i,v}-\hat{y}_{i,v}|,
    \qquad
    \mathrm{MSE}
    =
    \frac{1}{HV}
    \sum_{i=T+1}^{T+H}
    \sum_{v=1}^{V}
    (y_{i,v}-\hat{y}_{i,v})^2 ,
\end{equation}
where $H$ denotes the forecasting horizon, $V$ the number of variables,
$y_{i,v}$ the ground-truth value of variable $v$ at time step $i$, and
$\hat{y}_{i,v}$ the corresponding prediction.

The point-wise reverse Huber (BerHu) loss is defined as
\begin{equation}
    \mathcal{L}_{\mathrm{BerHu}}(e)
    =
    \begin{cases}
        |e|, & |e| \leq \delta, \\[1mm]
        \dfrac{e^2+\delta^2}{2\delta}, & |e| > \delta,
    \end{cases}
\end{equation}
where the transition threshold is set to
\begin{equation}
    \delta
    =
    \frac{1}{5}
    \max_{\substack{i=T+1,\ldots,T+H\\v=1,\ldots,V}}
    |y_{i,v}-\hat{y}_{i,v}|.
\end{equation}
We report the mean reverse Huber loss as
\begin{equation}
    \mathrm{BerHu}
    =
    \frac{1}{HV}
    \sum_{i=T+1}^{T+H}
    \sum_{v=1}^{V}
    \mathcal{L}_{\mathrm{BerHu}}
    \left(y_{i,v}-\hat{y}_{i,v}\right).
\end{equation}

\paragraph{Training and hyperparameter tuning details.}
In the following, we provide details to our training.
We used Adam as the optimizer and cosine annealing  an warm up phase of 5 Epochs. Training was performed between 30 to 70 epochs.

For hyperparameter tuning, we used Optuna \cite{Optuna} with Tree Parzen estimator \cite{TPE}, we used 8 - 30 trials for each prediction horizon; HPO is performed at the validation set. To provide full reproducibility also in the time series setting we provide all the details of the hyperparameter optimization in Tab.~\ref{tab:hpo_hyperparameters}.

\begin{table}[htbp]
    \centering
    \caption{Hyperparameter search spaces used for HPO. Deviations are indicated explicitly.}
    \label{tab:hpo_hyperparameters}
    \begin{tabular}{lll}
        \toprule
        \textbf{Hyperparameter} & \textbf{Default search space} & \textbf{Dataset-specific space} \\
        \midrule
        Learning rate ($lr$)
            & $[10^{-4}, 5 \times 10^{-3}]$
            & --- \\
        Forecasting heads
            & $\{1,2, 3\}$
            & ---  \\
        Amplitude scale ($s$)
            & $[0,0.5]$
            & --- \\

         Sequence length
            & $[520,720]$
            &---\\
        \midrule
        \multicolumn{3}{l}{\textit{Fixed parameters}} \\
        \midrule
        Batch size
            & $32$
            & --- \\
        Optimizer
            & Adam
            & --- \\
        Scheduler
            & Cosine annealing
            & --- \\
        Weight decay
            & $10^{-5}$
            & ---  \\
        Convolutional layers
            & $8$
            & --- \\
        Maximum kernel size
            & $160$
            & Traffic: 10 \\
        Minimum kernel size
            & $30$
            & Traffic: 3\\
        Number of epochs
            & $30$
            & Traffic: $70$ \\
        Loss 
            & MAE 
            & Traffic: reverse Huber \\
        \bottomrule
    \end{tabular}
\end{table}

\section{Full Forecasting Results}

Table~\ref{app:fullresults} reports the complete long-term forecasting results across all datasets and prediction horizons. Table~\ref{tab:win_counts} provides a compact overview of these results by counting how often each model ranks first or second across the evaluated settings.

\begin{table}[p] \label{app:fullresults}
\centering
\caption{Full multivariate long-term forecasting results across all datasets and prediction horizons. Lower is better for both MSE and MAE. The best result in each row is shown in \textbf{\textcolor{red}{bold red}}, the second-best is \underline{\textcolor{blue}{underlined blue}}. A dash (-) indicates the model was not evaluated on that dataset. A thin rule sets each dataset's \textit{Avg} row apart from its horizon rows.}
\label{tab:full_forecast_results}
\renewcommand{\arraystretch}{1.05}
\setlength{\tabcolsep}{2pt}
\resizebox{\linewidth}{!}{%
\scriptsize
\begin{tabular}{llcccccccccccccccccccccccccccc}
\toprule
Dataset & Horizon & \multicolumn{2}{c}{\rotatebox{60}{STeP-TS}} & \multicolumn{2}{c}{\rotatebox{60}{ModTCN}} & \multicolumn{2}{c}{\rotatebox{60}{xLSTM-}} & \multicolumn{2}{c}{\rotatebox{60}{xLSTMTime}} & \multicolumn{2}{c}{\rotatebox{60}{TimeMix}} & \multicolumn{2}{c}{\rotatebox{60}{TSMixer}} & \multicolumn{2}{c}{\rotatebox{60}{CycleNet}} & \multicolumn{2}{c}{\rotatebox{60}{DLinear}} & \multicolumn{2}{c}{\rotatebox{60}{TiDE}} & \multicolumn{2}{c}{\rotatebox{60}{S-Mamba}} & \multicolumn{2}{c}{\rotatebox{60}{PatchTST}} & \multicolumn{2}{c}{\rotatebox{60}{TimesNet}} & \multicolumn{2}{c}{\rotatebox{60}{Timer-XL}} & \multicolumn{2}{c}{\rotatebox{60}{MoiraiBase}} \\
\cmidrule(lr){3-4}\cmidrule(lr){5-6}\cmidrule(lr){7-8}\cmidrule(lr){9-10}\cmidrule(lr){11-12}\cmidrule(lr){13-14}\cmidrule(lr){15-16}\cmidrule(lr){17-18}\cmidrule(lr){19-20}\cmidrule(lr){21-22}\cmidrule(lr){23-24}\cmidrule(lr){25-26}\cmidrule(lr){27-28}\cmidrule(lr){29-30}
 & & MSE & MAE & MSE & MAE & MSE & MAE & MSE & MAE & MSE & MAE & MSE & MAE & MSE & MAE & MSE & MAE & MSE & MAE & MSE & MAE & MSE & MAE & MSE & MAE & MSE & MAE & MSE & MAE \\
\midrule
\multirow{5}{*}{Weather} & 96 & \textbf{\textcolor{red}{0.142}} & \textbf{\textcolor{red}{0.181}} & 0.149 & 0.200 & \underline{\textcolor{blue}{0.143}} & \underline{\textcolor{blue}{0.184}} & 0.144 & 0.187 & 0.147 & 0.197 & 0.145 & 0.198 & 0.148 & 0.200 & 0.176 & 0.237 & 0.166 & 0.222 & 0.165 & 0.210 & 0.149 & 0.198 & 0.172 & 0.220 & 0.171 & 0.225 & 0.220 & 0.217 \\
 & 192 & \textbf{\textcolor{red}{0.183}} & \textbf{\textcolor{red}{0.222}} & 0.196 & 0.245 & \underline{\textcolor{blue}{0.186}} & \underline{\textcolor{blue}{0.226}} & 0.192 & 0.236 & 0.189 & 0.239 & 0.191 & 0.242 & 0.190 & 0.240 & 0.220 & 0.282 & 0.209 & 0.263 & 0.214 & 0.252 & 0.194 & 0.241 & 0.219 & 0.261 & 0.221 & 0.271 & 0.271 & 0.259 \\
 & 336 & \textbf{\textcolor{red}{0.232}} & \textbf{\textcolor{red}{0.264}} & 0.238 & 0.277 & \underline{\textcolor{blue}{0.236}} & \underline{\textcolor{blue}{0.266}} & 0.237 & 0.272 & 0.241 & 0.280 & 0.242 & 0.280 & 0.242 & 0.283 & 0.265 & 0.319 & 0.254 & 0.301 & 0.274 & 0.297 & 0.306 & 0.282 & 0.280 & 0.306 & 0.274 & 0.311 & 0.286 & 0.297 \\
 & 720 & \textbf{\textcolor{red}{0.306}} & \textbf{\textcolor{red}{0.319}} & 0.314 & 0.334 & \underline{\textcolor{blue}{0.310}} & \underline{\textcolor{blue}{0.323}} & 0.313 & 0.326 & 0.310 & 0.330 & 0.320 & 0.336 & 0.312 & 0.333 & 0.323 & 0.362 & 0.313 & 0.340 & 0.350 & 0.345 & 0.314 & 0.334 & 0.365 & 0.359 & 0.356 & 0.370 & 0.373 & 0.354 \\
\cmidrule(lr){2-30}
 & \textit{Avg} & \textbf{\textcolor{red}{0.216}} & \textbf{\textcolor{red}{0.247}} & 0.224 & 0.264 & \underline{\textcolor{blue}{0.219}} & \underline{\textcolor{blue}{0.250}} & 0.222 & 0.255 & 0.222 & 0.262 & 0.225 & 0.264 & 0.223 & 0.264 & 0.246 & 0.300 & 0.236 & 0.282 & 0.251 & 0.276 & 0.241 & 0.264 & 0.259 & 0.287 & 0.256 & 0.294 & 0.288 & 0.282 \\
\midrule
\multirow{5}{*}{Electricity} & 96 & 0.130 & \underline{\textcolor{blue}{0.220}} & 0.129 & 0.226 & \textbf{\textcolor{red}{0.126}} & \textbf{\textcolor{red}{0.218}} & 0.128 & 0.221 & 0.129 & 0.224 & 0.131 & 0.229 & \underline{\textcolor{blue}{0.126}} & 0.221 & 0.140 & 0.237 & 0.132 & 0.229 & 0.139 & 0.235 & 0.129 & 0.222 & 0.168 & 0.272 & 0.141 & 0.237 & 0.160 & 0.250 \\
 & 192 & 0.146 & \underline{\textcolor{blue}{0.235}} & \underline{\textcolor{blue}{0.143}} & 0.239 & 0.144 & 0.235 & 0.150 & 0.243 & \textbf{\textcolor{red}{0.140}} & \textbf{\textcolor{red}{0.220}} & 0.151 & 0.246 & 0.144 & 0.237 & 0.153 & 0.249 & 0.147 & 0.243 & 0.159 & 0.255 & 0.147 & 0.240 & 0.184 & 0.289 & 0.159 & 0.254 & 0.175 & 0.263 \\
 & 336 & 0.162 & \underline{\textcolor{blue}{0.251}} & 0.161 & 0.259 & \textbf{\textcolor{red}{0.157}} & \textbf{\textcolor{red}{0.250}} & 0.166 & 0.259 & 0.161 & 0.255 & 0.161 & 0.261 & \underline{\textcolor{blue}{0.159}} & 0.255 & 0.169 & 0.267 & 0.161 & 0.261 & 0.176 & 0.272 & 0.163 & 0.259 & 0.198 & 0.300 & 0.177 & 0.272 & 0.187 & 0.277 \\
 & 720 & 0.200 & 0.283 & 0.191 & 0.286 & \textbf{\textcolor{red}{0.183}} & \textbf{\textcolor{red}{0.276}} & \underline{\textcolor{blue}{0.185}} & \underline{\textcolor{blue}{0.276}} & 0.194 & 0.287 & 0.197 & 0.293 & 0.196 & 0.290 & 0.203 & 0.301 & 0.196 & 0.294 & 0.204 & 0.298 & 0.197 & 0.290 & 0.220 & 0.320 & 0.219 & 0.308 & 0.228 & 0.309 \\
\cmidrule(lr){2-30}
 & \textit{Avg} & 0.160 & \underline{\textcolor{blue}{0.247}} & \underline{\textcolor{blue}{0.156}} & 0.253 & \textbf{\textcolor{red}{0.153}} & \textbf{\textcolor{red}{0.245}} & 0.157 & 0.250 & 0.156 & 0.247 & 0.160 & 0.257 & 0.156 & 0.251 & 0.166 & 0.264 & 0.159 & 0.257 & 0.170 & 0.265 & 0.159 & 0.253 & 0.193 & 0.295 & 0.174 & 0.268 & 0.188 & 0.275 \\
\midrule
\multirow{5}{*}{Traffic} & 96 & 0.388 & 0.262 & 0.368 & 0.253 & \underline{\textcolor{blue}{0.357}} & \textbf{\textcolor{red}{0.236}} & 0.358 & \underline{\textcolor{blue}{0.242}} & 0.360 & 0.249 & 0.376 & 0.264 & 0.374 & 0.268 & 0.410 & 0.282 & \textbf{\textcolor{red}{0.336}} & 0.253 & 0.382 & 0.261 & 0.360 & 0.249 & 0.593 & 0.321 & - & - & - & - \\
 & 192 & 0.398 & 0.264 & 0.379 & 0.261 & 0.377 & \textbf{\textcolor{red}{0.241}} & 0.378 & 0.253 & \underline{\textcolor{blue}{0.375}} & \underline{\textcolor{blue}{0.250}} & 0.397 & 0.277 & 0.390 & 0.275 & 0.423 & 0.287 & \textbf{\textcolor{red}{0.346}} & 0.257 & 0.396 & 0.267 & 0.379 & 0.256 & 0.617 & 0.336 & - & - & - & - \\
 & 336 & 0.418 & 0.268 & 0.397 & 0.270 & 0.394 & \textbf{\textcolor{red}{0.250}} & 0.392 & 0.261 & \underline{\textcolor{blue}{0.385}} & 0.270 & 0.413 & 0.290 & 0.405 & 0.282 & 0.436 & 0.296 & \textbf{\textcolor{red}{0.355}} & \underline{\textcolor{blue}{0.260}} & 0.417 & 0.276 & 0.392 & 0.264 & 0.629 & 0.336 & - & - & - & - \\
 & 720 & 0.453 & 0.289 & 0.440 & 0.296 & 0.439 & 0.283 & 0.434 & 0.287 & \underline{\textcolor{blue}{0.430}} & \underline{\textcolor{blue}{0.281}} & 0.444 & 0.306 & 0.441 & 0.302 & 0.466 & 0.315 & \textbf{\textcolor{red}{0.386}} & \textbf{\textcolor{red}{0.273}} & 0.460 & 0.300 & 0.432 & 0.286 & 0.640 & 0.350 & - & - & - & - \\
\cmidrule(lr){2-30}
 & \textit{Avg} & 0.414 & 0.271 & 0.396 & 0.270 & 0.392 & \textbf{\textcolor{red}{0.253}} & 0.391 & \underline{\textcolor{blue}{0.261}} & \underline{\textcolor{blue}{0.388}} & 0.263 & 0.408 & 0.284 & 0.403 & 0.282 & 0.434 & 0.295 & \textbf{\textcolor{red}{0.356}} & 0.261 & 0.414 & 0.276 & 0.391 & 0.264 & 0.620 & 0.336 & - & - & - & - \\
\midrule
\multirow{5}{*}{ETTh1} & 96 & 0.365 & \underline{\textcolor{blue}{0.389}} & 0.368 & 0.394 & \textbf{\textcolor{red}{0.359}} & \textbf{\textcolor{red}{0.386}} & 0.368 & 0.395 & \underline{\textcolor{blue}{0.361}} & 0.390 & 0.361 & 0.392 & 0.382 & 0.403 & 0.375 & 0.399 & 0.375 & 0.398 & 0.386 & 0.405 & 0.370 & 0.400 & 0.384 & 0.402 & 0.369 & 0.391 & 0.376 & 0.392 \\
 & 192 & \textbf{\textcolor{red}{0.396}} & \textbf{\textcolor{red}{0.409}} & 0.405 & \underline{\textcolor{blue}{0.413}} & 0.402 & 0.417 & \underline{\textcolor{blue}{0.401}} & 0.416 & 0.409 & 0.414 & 0.404 & 0.418 & 0.421 & 0.426 & 0.405 & 0.416 & 0.412 & 0.422 & 0.443 & 0.437 & 0.413 & 0.429 & 0.436 & 0.429 & 0.405 & 0.413 & 0.412 & 0.413 \\
 & 336 & 0.431 & 0.428 & \textbf{\textcolor{red}{0.391}} & \textbf{\textcolor{red}{0.412}} & \underline{\textcolor{blue}{0.408}} & 0.429 & 0.422 & 0.437 & 0.430 & 0.429 & 0.420 & 0.431 & 0.449 & 0.444 & 0.439 & 0.443 & 0.435 & 0.433 & 0.489 & 0.468 & 0.422 & 0.440 & 0.491 & 0.469 & 0.418 & \underline{\textcolor{blue}{0.423}} & 0.433 & 0.428 \\
 & 720 & 0.437 & 0.456 & 0.450 & 0.461 & \textbf{\textcolor{red}{0.419}} & 0.448 & 0.441 & 0.465 & 0.445 & 0.460 & 0.463 & 0.472 & 0.486 & 0.487 & 0.472 & 0.490 & 0.454 & 0.465 & 0.502 & 0.489 & 0.447 & 0.468 & 0.521 & 0.500 & \underline{\textcolor{blue}{0.423}} & \textbf{\textcolor{red}{0.441}} & 0.447 & \underline{\textcolor{blue}{0.444}} \\
\cmidrule(lr){2-30}
 & \textit{Avg} & 0.407 & 0.421 & \underline{\textcolor{blue}{0.404}} & 0.420 & \textbf{\textcolor{red}{0.397}} & 0.420 & 0.408 & 0.428 & 0.411 & 0.423 & 0.412 & 0.428 & 0.435 & 0.440 & 0.423 & 0.437 & 0.419 & 0.430 & 0.455 & 0.450 & 0.413 & 0.434 & 0.458 & 0.450 & 0.404 & \textbf{\textcolor{red}{0.417}} & 0.417 & \underline{\textcolor{blue}{0.419}} \\
\midrule
\multirow{5}{*}{ETTm1} & 96 & 0.289 & \underline{\textcolor{blue}{0.335}} & 0.292 & 0.346 & \textbf{\textcolor{red}{0.275}} & \textbf{\textcolor{red}{0.328}} & 0.286 & 0.335 & 0.291 & 0.340 & \underline{\textcolor{blue}{0.285}} & 0.339 & 0.297 & 0.351 & 0.299 & 0.343 & 0.306 & 0.349 & 0.333 & 0.368 & 0.293 & 0.346 & 0.338 & 0.375 & 0.317 & 0.356 & 0.363 & 0.356 \\
 & 192 & 0.337 & \underline{\textcolor{blue}{0.356}} & 0.332 & 0.368 & \textbf{\textcolor{red}{0.319}} & \textbf{\textcolor{red}{0.354}} & 0.329 & 0.361 & \underline{\textcolor{blue}{0.327}} & 0.365 & 0.327 & 0.365 & 0.338 & 0.377 & 0.335 & 0.365 & 0.335 & 0.366 & 0.376 & 0.390 & 0.333 & 0.370 & 0.374 & 0.387 & 0.358 & 0.381 & 0.388 & 0.375 \\
 & 336 & 0.364 & \underline{\textcolor{blue}{0.376}} & 0.365 & 0.391 & \textbf{\textcolor{red}{0.353}} & \textbf{\textcolor{red}{0.374}} & 0.358 & 0.379 & 0.360 & 0.381 & \underline{\textcolor{blue}{0.356}} & 0.382 & 0.374 & 0.400 & 0.369 & 0.386 & 0.364 & 0.384 & 0.408 & 0.413 & 0.369 & 0.392 & 0.410 & 0.411 & 0.386 & 0.401 & 0.416 & 0.392 \\
 & 720 & 0.420 & \underline{\textcolor{blue}{0.409}} & 0.416 & 0.417 & \textbf{\textcolor{red}{0.409}} & \textbf{\textcolor{red}{0.407}} & 0.416 & 0.411 & 0.415 & 0.417 & 0.419 & 0.414 & 0.431 & 0.425 & 0.425 & 0.421 & \underline{\textcolor{blue}{0.413}} & 0.413 & 0.475 & 0.448 & 0.416 & 0.420 & 0.478 & 0.450 & 0.430 & 0.431 & 0.460 & 0.418 \\
\cmidrule(lr){2-30}
 & \textit{Avg} & 0.353 & \underline{\textcolor{blue}{0.369}} & 0.351 & 0.381 & \textbf{\textcolor{red}{0.339}} & \textbf{\textcolor{red}{0.366}} & \underline{\textcolor{blue}{0.347}} & 0.372 & 0.348 & 0.376 & 0.347 & 0.375 & 0.360 & 0.388 & 0.357 & 0.379 & 0.355 & 0.378 & 0.398 & 0.405 & 0.353 & 0.382 & 0.400 & 0.406 & 0.373 & 0.392 & 0.407 & 0.385 \\
\midrule
\multirow{5}{*}{ETTm2} & 96 & \underline{\textcolor{blue}{0.158}} & \underline{\textcolor{blue}{0.245}} & 0.166 & 0.256 & \textbf{\textcolor{red}{0.157}} & \textbf{\textcolor{red}{0.244}} & 0.164 & 0.250 & 0.164 & 0.254 & 0.163 & 0.252 & 0.176 & 0.265 & 0.167 & 0.260 & 0.161 & 0.251 & 0.179 & 0.263 & 0.166 & 0.256 & 0.187 & 0.267 & 0.189 & 0.277 & 0.205 & 0.273 \\
 & 192 & \underline{\textcolor{blue}{0.214}} & \textbf{\textcolor{red}{0.282}} & 0.222 & 0.293 & \textbf{\textcolor{red}{0.213}} & \underline{\textcolor{blue}{0.285}} & 0.218 & 0.288 & 0.223 & 0.295 & 0.216 & 0.290 & 0.231 & 0.305 & 0.224 & 0.303 & 0.215 & 0.289 & 0.250 & 0.309 & 0.223 & 0.296 & 0.249 & 0.309 & 0.241 & 0.315 & 0.275 & 0.316 \\
 & 336 & 0.270 & \textbf{\textcolor{red}{0.322}} & 0.272 & 0.324 & 0.269 & \underline{\textcolor{blue}{0.322}} & 0.271 & 0.322 & 0.279 & 0.330 & \underline{\textcolor{blue}{0.268}} & 0.324 & 0.282 & 0.338 & 0.281 & 0.342 & \textbf{\textcolor{red}{0.267}} & 0.326 & 0.312 & 0.349 & 0.274 & 0.329 & 0.321 & 0.351 & 0.286 & 0.348 & 0.329 & 0.350 \\
 & 720 & 0.352 & \textbf{\textcolor{red}{0.375}} & \textbf{\textcolor{red}{0.351}} & 0.381 & \underline{\textcolor{blue}{0.351}} & \underline{\textcolor{blue}{0.377}} & 0.361 & 0.380 & 0.359 & 0.383 & 0.420 & 0.422 & 0.361 & 0.388 & 0.397 & 0.421 & 0.352 & 0.383 & 0.411 & 0.406 & 0.362 & 0.385 & 0.408 & 0.403 & 0.375 & 0.402 & 0.437 & 0.411 \\
\cmidrule(lr){2-30}
 & \textit{Avg} & \underline{\textcolor{blue}{0.249}} & \textbf{\textcolor{red}{0.306}} & 0.253 & 0.314 & \textbf{\textcolor{red}{0.248}} & \underline{\textcolor{blue}{0.307}} & 0.254 & 0.310 & 0.256 & 0.316 & 0.267 & 0.322 & 0.263 & 0.324 & 0.267 & 0.332 & 0.249 & 0.312 & 0.288 & 0.332 & 0.256 & 0.317 & 0.291 & 0.333 & 0.273 & 0.336 & 0.312 & 0.338 \\
\midrule
\multirow{5}{*}{ETTh2} & 96 & 0.272 & 0.332 & \textbf{\textcolor{red}{0.263}} & 0.332 & \underline{\textcolor{blue}{0.267}} & \underline{\textcolor{blue}{0.329}} & 0.273 & 0.333 & 0.271 & 0.330 & 0.274 & 0.341 & 0.352 & \textbf{\textcolor{red}{0.289}} & 0.289 & 0.353 & 0.270 & 0.336 & 0.296 & 0.348 & 0.274 & 0.337 & 0.340 & 0.374 & 0.283 & 0.342 & 0.294 & 0.330 \\
 & 192 & 0.335 & \underline{\textcolor{blue}{0.375}} & 0.320 & \textbf{\textcolor{red}{0.374}} & 0.338 & 0.375 & 0.340 & 0.378 & \underline{\textcolor{blue}{0.317}} & 0.402 & 0.339 & 0.385 & 0.395 & 0.383 & 0.383 & 0.418 & 0.332 & 0.380 & 0.376 & 0.396 & \textbf{\textcolor{red}{0.314}} & 0.382 & 0.402 & 0.414 & 0.340 & 0.379 & 0.365 & 0.375 \\
 & 336 & 0.366 & 0.403 & \textbf{\textcolor{red}{0.313}} & \textbf{\textcolor{red}{0.376}} & 0.367 & 0.401 & 0.373 & 0.403 & 0.332 & 0.396 & 0.361 & 0.406 & 0.423 & 0.448 & 0.448 & 0.465 & 0.360 & 0.407 & 0.424 & 0.431 & \underline{\textcolor{blue}{0.329}} & \underline{\textcolor{blue}{0.384}} & 0.452 & 0.452 & 0.366 & 0.400 & 0.376 & 0.390 \\
 & 720 & 0.387 & 0.425 & 0.392 & 0.433 & 0.388 & 0.424 & 0.398 & 0.430 & \textbf{\textcolor{red}{0.342}} & \textbf{\textcolor{red}{0.408}} & 0.445 & 0.470 & 0.451 & 0.605 & 0.605 & 0.551 & 0.419 & 0.451 & 0.426 & 0.444 & \underline{\textcolor{blue}{0.379}} & \underline{\textcolor{blue}{0.422}} & 0.462 & 0.468 & 0.397 & 0.431 & 0.416 & 0.433 \\
\cmidrule(lr){2-30}
 & \textit{Avg} & 0.340 & 0.384 & \underline{\textcolor{blue}{0.322}} & \textbf{\textcolor{red}{0.379}} & 0.340 & 0.382 & 0.346 & 0.386 & \textbf{\textcolor{red}{0.316}} & 0.384 & 0.355 & 0.401 & 0.405 & 0.431 & 0.431 & 0.447 & 0.345 & 0.394 & 0.381 & 0.405 & 0.324 & \underline{\textcolor{blue}{0.381}} & 0.414 & 0.427 & 0.347 & 0.388 & 0.363 & 0.382 \\
\bottomrule
\end{tabular}%
}
\end{table}

\begin{table}[p]
\centering
\caption{Number of times each model achieves the best (\textbf{bold red}) or second-best (\underline{underlined blue}) result in Table~\ref{tab:full_forecast_results}, broken down by metric. Counted over all 35 (dataset, horizon) rows, including the per-dataset averages. ``Best (total)'' sums the MSE and MAE best-counts; ``2nd (total)'' sums the second-best counts.}
\label{tab:win_counts}
\resizebox{0.8\linewidth}{!}{%
\small
\begin{tabular}{lcccccc}
\toprule
Model & MSE best & MSE 2nd & MAE best & MAE 2nd & Best (total) & 2nd (total) \\
\midrule
xLSTMix. & 15 & 9 & 15 & 10 & \textbf{30} & 19 \\
STeP-TS & 6 & 3 & 10 & 12 & \textbf{16} & 15 \\
ModTCN & 4 & 4 & 4 & 1 & \textbf{8} & 5 \\
TiDE & 6 & 1 & 1 & 1 & \textbf{7} & 2 \\
TimeMix & 3 & 7 & 2 & 2 & \textbf{5} & 9 \\
Timer-XL & 0 & 1 & 2 & 1 & \textbf{2} & 2 \\
PatchTST & 1 & 2 & 0 & 3 & \textbf{1} & 5 \\
CycleNet & 0 & 2 & 1 & 0 & \textbf{1} & 2 \\
xLSTMTime & 0 & 3 & 0 & 3 & \textbf{0} & 6 \\
TSMixer & 0 & 3 & 0 & 0 & \textbf{0} & 3 \\
MoiraiBase & 0 & 0 & 0 & 2 & \textbf{0} & 2 \\
DLinear & 0 & 0 & 0 & 0 & \textbf{0} & 0 \\
S-Mamba & 0 & 0 & 0 & 0 & \textbf{0} & 0 \\
TimesNet & 0 & 0 & 0 & 0 & \textbf{0} & 0 \\
\bottomrule
\end{tabular}%
}
\end{table}
\clearpage

\section{Pruning and Efficiency} \label{app:prune and efficency}
This section provides additional results on the computational efficiency of STeP-TS and the effect of structured channel pruning. We first compare computational cost across models and lookback lengths in Table~\ref{tab:pruning-summary}. We then evaluate the cost–accuracy trade-off of pruning, with Table~\ref{tab:effi} on the Weather dataset. Finally, Table~\ref{detailed: pruning} provides the complete per-dataset and per-horizon results.

In more detail, Table~\ref{detailed: pruning} reports the average computational improvements across datasets. We compare the pruning variant $p_c^{\mathrm{e}}$, which uses learnable aggregation weights $A_{v,c}$, against the corresponding variant without $A_{v,c}$.
For each dataset and forecasting horizon, we conduct three runs with different random seeds. As described in~\ref{pruning_schedule}, we perform iterative pruning after a warm-up period of 8 epochs. The first pruning step is performed after the warm-up, followed by subsequent pruning steps at intervals separated by a cooldown period of $E_r=5$ epochs. Within each pruning interval, we select the epoch with the best validation loss. Training lasts for a total of 40 epochs, with no further pruning allowed during the final 5 epochs. To ensure a fair comparison of the computational measurements on the same GPU, we perform a control run immediately before each pruning run using the corresponding seed. We note that the number of channels that can be pruned varies across seeds. This variation is expected, as the kernel size determines the lowest frequency that can be captured, while higher-frequency components remain less constrained. Consequently, the temporal information represented by individual channels can depend on the initialization, leading to differences in the number of channels identified for pruning across seeds. We provide more details on the speed-up measurements in \ref{app:speed-Up}.

\begin{table}[htb] 
\centering
\caption{Cost--accuracy trade-off of the two channel-pruning criteria, averaged over 20 dataset/horizon pairs. \emph{Dense} is the same with pruning disabled, trained from the same config, seeds, code and GPU in the same job and scored on the same epoch window. Latency is per sample at batch~32 on an L40S.}
\label{tab:pruning-summary}
\begin{tabular}{lrrrrrr}
\toprule
Criterion & Avg. Active Channels & FLOPs & Latency & Speedup & $\Delta$MSE\% & $\Delta$MAE\% \\
 & (of 8) & (rel.) & [$\mu$s/sample] & & & \\
\midrule
Dense & 8.00 & 1.00 & 49.0 & 1.00 & --- & --- \\
with  $A_{v,c}$ & 5.00 & 0.63 & 42.1 & \textbf{1.20} & +0.03 & -0.01 \\
w/o  $A_{v,c}$ & 5.97 & 0.75 & 45.4 & 1.08 & 0.00 & 0.00 \\
\bottomrule
\end{tabular}
\end{table}

As shown in table ~\ref{tab:pruning-summary},  pruning without $A_{v,c}$ has no measurable impact on performance, with an average degradation of $0\%$. In contrast, pruning with $A_{v,c}$ provides greater computational gains at the cost of a slightly higher, but still negligible, accuracy degradation of $0.03\%$ in MSE, while MAE even improves marginally by $0.01\%$.

Additionally,in Table ~\ref{tab:energy_vs_random_channels}, , we compare pruning with $A_{v,c}$ against randomly selecting channels for pruning across datasets. Interestingly, when the resulting pruned architectures are retrained from scratch, both approaches lead to a similar average performance degradation of approximately $0.3\%$. This suggests that training the pruned architecture from scratch is not a suitable strategy for preserving the performance of the original model, regardless of how the channels are selected. In contrast, when pruning is performed iteratively while retaining and fine-tuning the learned model, $A_{v,c}$ results in only a negligible performance degradation of $0.03\%$ and outperforms random pruning $0.08\%$. This advantage becomes even more pronounced as more channels are removed, suggesting that the model can compensate for the removal of a small number of channels, whereas more aggressive pruning requires a systematic selection of channels to avoid substantial performance degradation.
\vspace{3cm}

\begin{table}[htb]
\centering
\small
\setlength{\tabcolsep}{4.5pt}
\caption{$A_{v,c}$ versus random channel pruning, grouped by the number of scale channels removed per seed.
Entries are the change in test MSE relative to the dense control of the same seed, in percent;
the difference column is random minus energy in percentage points ($\pm$ one standard error over cells).
\emph{Pruned} is the model during iterative training and scored after the last removal;
\emph{retrained} is a fresh model trained from scratch with the final channel set fixed from epoch~0.
Random pruning is averaged over three independent draws. Cells are (dataset, horizon, seed) triples over five datasets, so cells of one combination are not independent.}
\label{tab:energy_vs_random_channels}
\resizebox{\textwidth}{!}{%
\begin{tabular}{rcc ccc ccc}
\toprule
 & & & \multicolumn{3}{c}{Pruned during training} & \multicolumn{3}{c}{Retrained from scratch} \\
\cmidrule(lr){4-6}\cmidrule(lr){7-9}
Removed & Cells & Combos &$A_{v,c}$ & Random & Random $-$ $A_{v,c}$ &$A_{v,c}$ & Random & Random $-$ $A_{v,c}$ \\
\midrule
0 &  4 &  4 & $-0.03$ & $-0.01$ & $+0.02 \pm 0.02$ & $-0.03$ & $-0.01$ & $+0.02 \pm 0.02$ \\
1 &  5 &  4 & $+0.01$ & $+0.02$ & $+0.01 \pm 0.09$ & $+0.04$ & $-0.15$ & $-0.19 \pm 0.26$ \\
2 & 15 & 11 & $-0.07$ & $-0.15$ & $-0.08 \pm 0.10$ & $+0.45$ & $+0.05$ & $-0.40 \pm 0.14$ \\
3 & 13 & 11 & $+0.06$ & $+0.11$ & $+0.05 \pm 0.03$ & $+0.44$ & $+0.46$ & $+0.02 \pm 0.19$ \\
4 & 12 & 10 & $-0.14$ & $-0.06$ & $+0.08 \pm 0.09$ & $-0.32$ & $-0.07$ & $+0.25 \pm 0.40$ \\
5 &  8 &  8 & $-0.05$ & $+0.26$ & $+0.31 \pm 0.29$ & $+0.04$ & $+0.56$ & $+0.52 \pm 0.60$ \\
6 &  3 &  2 & $+1.35$ & $+1.89$ & $+0.54 \pm 0.26$ & $+3.19$ & $+2.78$ & $-0.41 \pm 0.43$ \\
\midrule
All & 60 & 20 & $+0.03$ & $+0.10$ & $+0.08 \pm 0.05$ & $+0.31$ & $+0.30$ & $-0.01 \pm 0.13$ \\
\bottomrule
\end{tabular}%
}
\end{table}

\begin{table}
\caption{Per-horizon pruning results. \emph{Chan.} is the mean number of surviving scale channels of 8; FLOPs, latency and speedup are relative to that run's own dense pass. $\Delta$ is the change against dense on the same three seeds; \textbf{bold} marks the MSE changes that clear a paired two-sided $t$-test ($df=2$, $\alpha=0.05$) --- every other $\Delta$ lies inside the seed spread and should be read as no measurable change. The \emph{avg} line under each dataset averages its four horizons.} \label{detailed: pruning}
\scriptsize
\setlength{\tabcolsep}{2.5pt}
\begin{tabular}{ll rrr rrrrrr rrrrrr}
\toprule
& & \multicolumn{3}{c}{Dense} & \multicolumn{6}{c}{with $A_{v,c}$} & \multicolumn{6}{c}{w/o $A_{v,c}$} \\
\cmidrule(lr){3-5}\cmidrule(lr){6-11}\cmidrule(lr){12-17}
Dataset & $H$ & MSE & MAE & Lat. & Chan. & FLOPs & Lat. & Spd. & $\Delta$MSE\% & $\Delta$MAE\% & Chan. & FLOPs & Lat. & Spd. & $\Delta$MSE\% & $\Delta$MAE\% \\
 & & & & [$\mu$s] & (of 8) & (rel.) & [$\mu$s] & & & & (of 8) & (rel.) & [$\mu$s] & & & \\
\midrule
ETTh1 & 96 & 0.367 & 0.390 & 49.3 & 7.00 & 0.87 & 49.2 & 1.00 & +0.07 & -0.09 & 7.33 & 0.91 & 49.8 & 0.97 & +0.10 & -0.01 \\
 & 192 & 0.394 & 0.408 & 48.3 & 6.67 & 0.83 & 46.5 & 1.04 & +0.24 & +0.03 & 7.33 & 0.93 & 47.1 & 1.01 & +0.53 & +0.24 \\
 & 336 & 0.427 & 0.427 & 48.5 & 7.00 & 0.85 & 48.2 & 1.01 & +0.06 & +0.09 & 6.67 & 0.82 & 47.0 & 1.03 & +0.26 & +0.13 \\
 & 720 & 0.437 & 0.457 & 48.1 & 7.33 & 0.90 & 49.6 & 0.98 & -0.09 & -0.03 & 6.67 & 0.80 & 45.2 & 1.03 & 0.00 & +0.06 \\
\cmidrule(l){2-17}
 & \emph{avg} & 0.406 & 0.420 & 48.5 & 7.00 & 0.86 & 48.4 & 1.01 & +0.07 & 0.00 & 7.00 & 0.86 & 47.3 & 1.01 & +0.22 & +0.11 \\
\addlinespace[2pt]
ETTh2 & 96 & 0.273 & 0.332 & 48.2 & 5.67 & 0.72 & 43.5 & 1.10 & -0.71 & -0.51 & 7.00 & 0.87 & 47.1 & 1.00 & -0.12 & -0.11 \\
 & 192 & 0.332 & 0.373 & 48.1 & 3.67 & 0.47 & 34.3 & 1.38 & -0.10 & -0.09 & 6.67 & 0.84 & 45.1 & 1.04 & -0.43 & -0.28 \\
 & 336 & 0.387 & 0.411 & 49.5 & 4.00 & 0.51 & 37.1 & 1.32 & -0.76 & -0.38 & 6.00 & 0.76 & 45.0 & 1.07 & -0.59 & -0.15 \\
 & 720 & 0.384 & 0.424 & 49.0 & 4.00 & 0.51 & 36.8 & 1.32 & -0.10 & -0.12 & 6.00 & 0.74 & 44.8 & 1.06 & -0.06 & -0.11 \\
\cmidrule(l){2-17}
 & \emph{avg} & 0.344 & 0.385 & 48.7 & 4.33 & 0.55 & 37.9 & 1.28 & -0.42 & -0.27 & 6.42 & 0.80 & 45.5 & 1.04 & -0.30 & -0.16 \\
\addlinespace[2pt]
ETTm1 & 96 & 0.288 & 0.334 & 48.4 & 2.33 & 0.28 & 29.5 & 1.58 & +1.26 & +0.28 & 4.67 & 0.60 & 38.6 & 1.22 & +0.01 & +0.02 \\
 & 192 & 0.337 & 0.356 & 49.8 & 4.67 & 0.62 & 40.6 & 1.22 & +0.05 & +0.03 & 5.00 & 0.64 & 40.7 & 1.18 & +0.06 & +0.09 \\
 & 336 & 0.363 & 0.376 & 49.6 & 5.33 & 0.68 & 43.8 & 1.12 & 0.00 & +0.01 & 5.33 & 0.67 & 42.8 & 1.12 & -0.05 & -0.01 \\
 & 720 & 0.420 & 0.409 & 49.5 & 5.67 & 0.72 & 45.2 & 1.09 & -0.01 & +0.04 & 6.00 & 0.76 & 45.2 & 1.07 & +0.06 & +0.18 \\
\cmidrule(l){2-17}
 & \emph{avg} & 0.352 & 0.369 & 49.3 & 4.50 & 0.58 & 39.8 & 1.25 & +0.33 & +0.09 & 5.25 & 0.67 & 41.8 & 1.15 & +0.02 & +0.07 \\
\addlinespace[2pt]
ETTm2 & 96 & 0.158 & 0.245 & 49.3 & 5.33 & 0.68 & 42.7 & 1.15 & +0.09 & +0.13 & 5.67 & 0.73 & 43.7 & 1.09 & +0.03 & +0.06 \\
 & 192 & 0.214 & 0.283 & 49.3 & 4.67 & 0.61 & 40.5 & 1.21 & -0.16 & -0.14 & 5.67 & 0.73 & 43.9 & 1.09 & -0.07 & -0.15 \\
 & 336 & 0.269 & 0.320 & 48.4 & 4.00 & 0.51 & 36.3 & 1.33 & -0.26 & -0.03 & 5.67 & 0.75 & 42.4 & 1.11 & +0.01 & +0.03 \\
 & 720 & 0.351 & 0.375 & 48.1 & 5.00 & 0.65 & 41.0 & 1.17 & -0.01 & -0.03 & 5.33 & 0.68 & 41.0 & 1.14 & -0.08 & -0.04 \\
\cmidrule(l){2-17}
 & \emph{avg} & 0.248 & 0.306 & 48.8 & 4.75 & 0.61 & 40.1 & 1.21 & -0.09 & -0.02 & 5.58 & 0.72 & 42.7 & 1.11 & -0.03 & -0.03 \\
\addlinespace[2pt]
Weather & 96 & 0.142 & 0.181 & 49.2 & 4.00 & 0.49 & 41.3 & 1.32 & +0.39 & +0.12 & 5.67 & 0.71 & 49.9 & 1.08 & +0.11 & +0.03 \\
 & 192 & 0.182 & 0.222 & 48.8 & 3.00 & 0.39 & 36.8 & 1.42 & \textbf{+0.50} & +0.42 & 5.33 & 0.66 & 48.0 & 1.12 & +0.15 & +0.11 \\
 & 336 & 0.232 & 0.263 & 49.7 & 5.33 & 0.70 & 49.3 & 1.12 & +0.07 & +0.06 & 6.00 & 0.74 & 52.2 & 1.05 & +0.03 & -0.04 \\
 & 720 & 0.306 & 0.319 & 50.7 & 5.33 & 0.69 & 49.6 & 1.10 & \textbf{+0.02} & +0.03 & 5.33 & 0.69 & 49.5 & 1.11 & +0.01 & +0.03 \\
\cmidrule(l){2-17}
 & \emph{avg} & 0.216 & 0.246 & 49.6 & 4.42 & 0.56 & 44.3 & 1.24 & +0.25 & +0.16 & 5.58 & 0.70 & 49.9 & 1.09 & +0.08 & +0.03 \\
\midrule
\textbf{Mean} & & 0.313 & 0.345 & 49.0 & 5.00 & \textbf{0.63} & 42.1 & \textbf{1.20} & +0.03 & -0.01 & 5.97 & 0.75 & 45.4 & 1.08 & 0.00 & 0.00 \\
\bottomrule
\end{tabular}
\end{table}

\begin{table}[!ht]
\centering
\small
\setlength{\tabcolsep}{4pt}
\caption{Computational cost on Weather at (H=336), measured at three matched lookbacks on one RTX 3080 Ti with batch size 32. Parameters are exact counts; FLOPs are \texttt{thop} estimates from a batch-1 pass. Latency, throughput, training-step time, and peak memory are averaged over repeated runs after warm-up. STeP-TS uses the standard HPO configuration; pruning removes 3 of 8 scale channels on average, with all other settings unchanged. ($^{\ddagger}$)PatchTST uses default hyperparameters. Batch size is fixed at 32 for comparability, so peak training memory may differ from the baseline authors' settings.}\label{tab:effi}
\scriptsize

\begin{tabular}{llrrrrrrr}
\toprule
$L$ & Model & \multicolumn{1}{c}{Params} & \multicolumn{1}{c}{FLOPs}
& \multicolumn{1}{c}{Latency} & \multicolumn{1}{c}{Throughput}
& \multicolumn{1}{c}{Train step}
& \multicolumn{1}{c}{Mem$_{\text{inf}}$} & \multicolumn{1}{c}{Mem$_{\text{train}}$} \\
& & \multicolumn{1}{c}{[K]} & \multicolumn{1}{c}{[M]}
& \multicolumn{1}{c}{[$\mu$s/sample]}
& \multicolumn{1}{c}{[samples/s]} & \multicolumn{1}{c}{[ms]}
& \multicolumn{1}{c}{[MB]} & \multicolumn{1}{c}{[B]} \\
\midrule
\multirow{8}{*}{336} & STeP-TS, 8 ch, linear head & 128.9 & 9.1 & 61.5 & 16,260 & 8.5 & 9.7 & 51.4 \\
 & STeP-TS, 8 ch, 1 head & 128.9 & 9.1 & 71.9 & 13,916 & 8.5 & 17.9 & 51.4 \\
 & STeP-TS, 8 ch, 2 heads & 242.2 & 13.9 & 73.1 & 13,685 & 8.6 & 18.3 & 53.1 \\
 & STeP-TS, 8 ch, 3 heads$^{\dagger}$ & 355.4 & 18.6 & 73.5 & 13,608 & 8.8 & 18.8 & 54.9 \\
 & STeP-TS, 5 ch, 3 heads (pruned) & 355.1 & 17.0 & 55.0 & 18,193 & 6.3 & 18.8 & 43.2 \\
 & xLSTM-Mixer & 961.1 & 41.5 & 84.2 & 11,882 & 11.2 & 21.2 & 61.6 \\
 & ModernTCN & 3,790.8 & 401.0 & 121.4 & 8,238 & 18.7 & 39.1 & 807.4 \\
 & PatchTST$^{\ddagger}$ & 2,206.4 & 779.4 & 344.4 & 2,904 & 36.0 & 36.5 & 1,227.2 \\
\midrule
\multirow{8}{*}{512} & STeP-TS, 8 ch, linear head & 195.4 & 13.9 & 66.5 & 15,046 & 10.7 & 18.2 & 67.5 \\
 & STeP-TS, 8 ch, 1 head & 195.5 & 13.9 & 70.0 & 14,276 & 10.6 & 18.2 & 67.5 \\
 & STeP-TS, 8 ch, 2 heads & 367.9 & 21.1 & 76.7 & 13,034 & 10.9 & 18.9 & 69.8 \\
 & STeP-TS, 8 ch, 3 heads$^{\dagger}$ & 540.3 & 28.3 & 72.2 & 13,840 & 11.1 & 19.5 & 72.4 \\
 & STeP-TS, 5 ch, 3 heads (pruned) & 539.9 & 26.0 & 57.4 & 17,420 & 7.8 & 19.5 & 55.7 \\
 & xLSTM-Mixer & 1,138.5 & 44.0 & 83.9 & 11,913 & 8.8 & 21.9 & 63.0 \\
 & ModernTCN & 4,736.9 & 611.1 & 682.5 & 1,465 & 43.2 & 45.6 & 1,209.0 \\
 & PatchTST$^{\ddagger}$ & 3,152.6 & 1,187.7 & 475.5 & 2,103 & 50.0 & 50.0 & 2,202.3 \\
\midrule
\multirow{8}{*}{768} & STeP-TS, 8 ch, linear head & 292.3 & 20.8 & 68.0 & 14,700 & 13.2 & 19.1 & 92.0 \\
 & STeP-TS, 8 ch, 1 head & 292.4 & 20.8 & 71.2 & 14,052 & 13.6 & 19.1 & 92.0 \\
 & STeP-TS, 8 ch, 2 heads & 550.8 & 31.7 & 73.8 & 13,544 & 13.9 & 20.1 & 95.0 \\
 & STeP-TS, 8 ch, 3 heads$^{\dagger}$ & 809.2 & 42.5 & 73.6 & 13,586 & 14.0 & 21.1 & 98.0 \\
 & STeP-TS, 5 ch, 3 heads (pruned) & 808.8 & 38.9 & 55.1 & 18,150 & 9.5 & 20.6 & 73.5 \\
 & xLSTM-Mixer & 1,396.5 & 47.6 & 85.8 & 11,654 & 7.3 & 22.9 & 68.6 \\
 & ModernTCN & 6,113.2 & 916.6 & 1018.1 & 982 & 63.2 & 56.8 & 1,788.8 \\
 & PatchTST$^{\ddagger}$ & 4,528.8 & 1,781.6 & 771.3 & 1,296 & 81.2 & 77.3 & 4,296.8 \\
\bottomrule
\end{tabular}
\end{table}

\newpage

\end{document}